\documentclass[10pt,twocolumn,letterpaper]{article}

\usepackage{iccv}
\usepackage{times}
\usepackage{epsfig}
\usepackage{graphicx}
\usepackage{amsmath}
\usepackage{amssymb}
\usepackage{multirow}
\usepackage{booktabs}

\def\eg{{\em e.g.}}
\def\ie{{\em i.e.}}
\def\etal{{\em et al.}}

\usepackage[pagebackref=true,breaklinks=true,letterpaper=true,colorlinks,bookmarks=false]{hyperref}
\iccvfinalcopy %

\def\iccvPaperID{4121} 

\ificcvfinal\fi

\begin{document}

\title{Towards Real-World Prohibited Item Detection: A Large-Scale X-ray Benchmark}

\author{
	Boying Wang$^{1,2}$,
	Libo Zhang$^{1,2,3}$\thanks{Corresponding author (libo@iscas.ac.cn). This work was supported by the Key Research Program of Frontier Sciences, CAS, Grant No. ZDBS-LY-JSC038, the National Natural Science Foundation of China, Grant No. 61807033. Libo Zhang was supported by Youth Innovation Promotion Association, CAS (2020111), and Outstanding Youth Scientist Project of ISCAS. The PIDray dataset are available at \url{https://github.com/bywang2018/security-dataset}.},
	Longyin Wen$^{4}$, Xianglong Liu$^{5}$, Yanjun Wu$^{1}$ \\
	 $^{1}$State Key Laboratory of Computer Science, ISCAS, Beijing, China \\ 
	 $^{2}$University of Chinese Academy of Sciences, Beijing, China.  \\ 
   $^{3}$Hangzhou Institute for Advanced Study, UCAS, Hangzhou, China. \\
	 $^{4}$JD Finance America Corporation, Mountain View, CA, USA. \\
	 $^{5}$ Beihang University, Beijing, China. \\
   \tt\small \{boying2018, libo, yanjun\}@iscas.ac.cn
   \tt\small longyin.wen.cv@gmail.com
   \tt\small xlliu@nlsde.buaa.edu.cn
}

\maketitle
\ificcvfinal\thispagestyle{empty}\fi

\begin{abstract}
Automatic security inspection using computer vision technology is a challenging task in real-world scenarios due to various factors, including intra-class variance, class imbalance, and occlusion. Most of the previous methods rarely solve the cases that the prohibited items are deliberately hidden in messy objects due to the lack of large-scale datasets, restricted their applications in real-world scenarios. Towards real-world prohibited item detection, we collect a large-scale dataset, named as PIDray, which covers various cases in real-world scenarios for prohibited item detection, especially for deliberately hidden items. With an intensive amount of effort, our dataset contains $12$ categories of prohibited items in $47,677$ X-ray images with high-quality annotated segmentation masks and bounding boxes. To the best of our knowledge, it is the largest prohibited items detection dataset to date. Meanwhile, we design the selective dense attention network (SDANet) to construct a strong baseline, which consists of the dense attention module and the dependency refinement module. The dense attention module formed by the spatial and channel-wise dense attentions, is designed to learn the discriminative features to boost the performance. The dependency refinement module is used to exploit the dependencies of multi-scale features. Extensive experiments conducted on the collected PIDray dataset demonstrate that the proposed method performs favorably against the state-of-the-art methods, especially for detecting the deliberately hidden items.
\end{abstract}
\section{Introduction}
Security inspection is a process of checking assets against set criteria and the evaluation of security systems and access controls to ensure safety, which is important to uncover any potential risks in various scenarios, such as public transportation and sensitive departments. In practice, the inspectors are required to monitor the scanned X-ray images acquired by the security inspection machine to uncover prohibited items, such as guns, ammunition, explosives, corrosive substances, toxic and radioactive substances. However, the inspectors struggle to localize prohibited items hidden in messy objects accurately and efficiently, which poses a great threat to safety. 

\begin{figure}[t]
\centering
\includegraphics[width=0.9\linewidth]{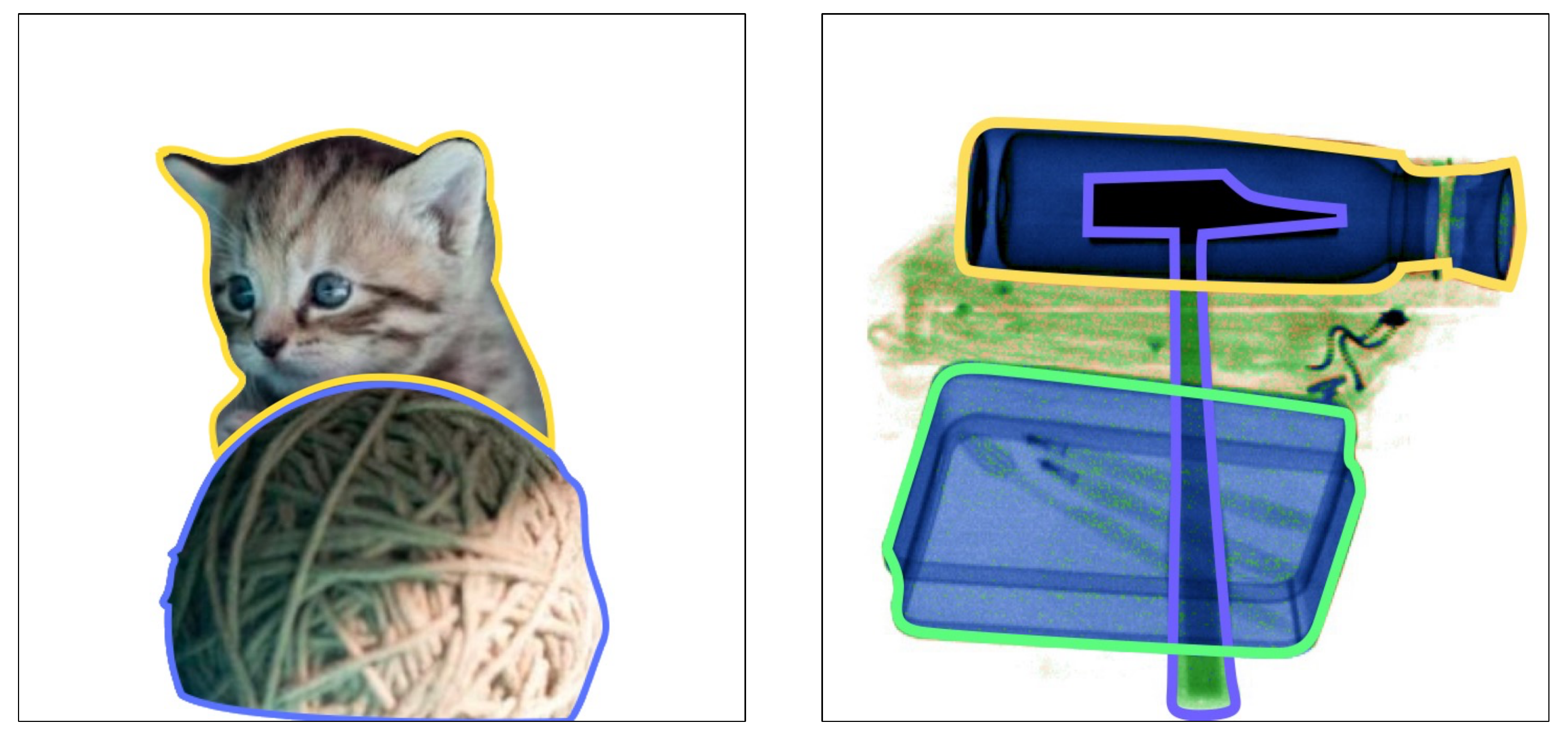}
\caption{Comparisons between the natural image (left) and X-ray image (right).}
\label{fig:overlapping}
\end{figure}

In recent years, due to the substantial development of deep learning and computer vision technologies\cite{ren2015faster, liu2016ssd, tian2019fcos, DBLP:conf/cvpr/JiWZDWZLH20, DBLP:conf/eccv/JiDZWWZHL20, DBLP:conf/eccv/LiDZWLWZ20, DBLP:conf/mm/CaiDZWWWL20}, automatic security inspection of prohibited items becomes possible. The security inspectors can quickly identify the locations and categories of prohibited items relying on computer vision technology. Most of the previous object detection algorithms in computer vision are designed to detect objects in natural images, which are not optimal for detection in X-ray images. In addition, X-rays have strong penetrating power, different materials in the object absorb X-rays to different degrees, resulting in different colors. Meanwhile, the contours of the occluder and the occluded objects in the x-ray are mixed together. As shown in Figure \ref{fig:overlapping}, compared with natural images, X-ray images have a quite different appearance and edges of objects and background, which brings new challenges in appearance modeling for X-ray detection. To advance the developments of prohibited items detection in X-ray images, some recent attempts devote to construct security inspection benchmarks \cite{mery2015gdxray,akcay2017evaluation,akcay2018using,miao2019sixray,wei2020occluded}. However, most of them fail to meet the requirements in real-world applications for three reasons. (1) Existing datasets only contain a small number and very few categories of prohibited items (\eg, \textit{knife}, \textit{gun} and \textit{scissors}). For example, some common prohibited items such as \textit{powerbank}, \textit{lighter} and \textit{sprayer} are not included. (2) Some real-world scenarios require high security level based on accurate predictions of masks and categories of prohibited items. The image-level or bounding box-level annotations in previous datasets are not sufficient to train algorithms in such scenarios. (3) Detecting prohibited items hidden in messy objects is one of the most significant challenges in security inspection. Nevertheless, few studies are developed towards this goal due to the lack of comprehensive datasets covering such cases. 

\begin{figure}[t]
\centering
\includegraphics[width=1.0\linewidth]{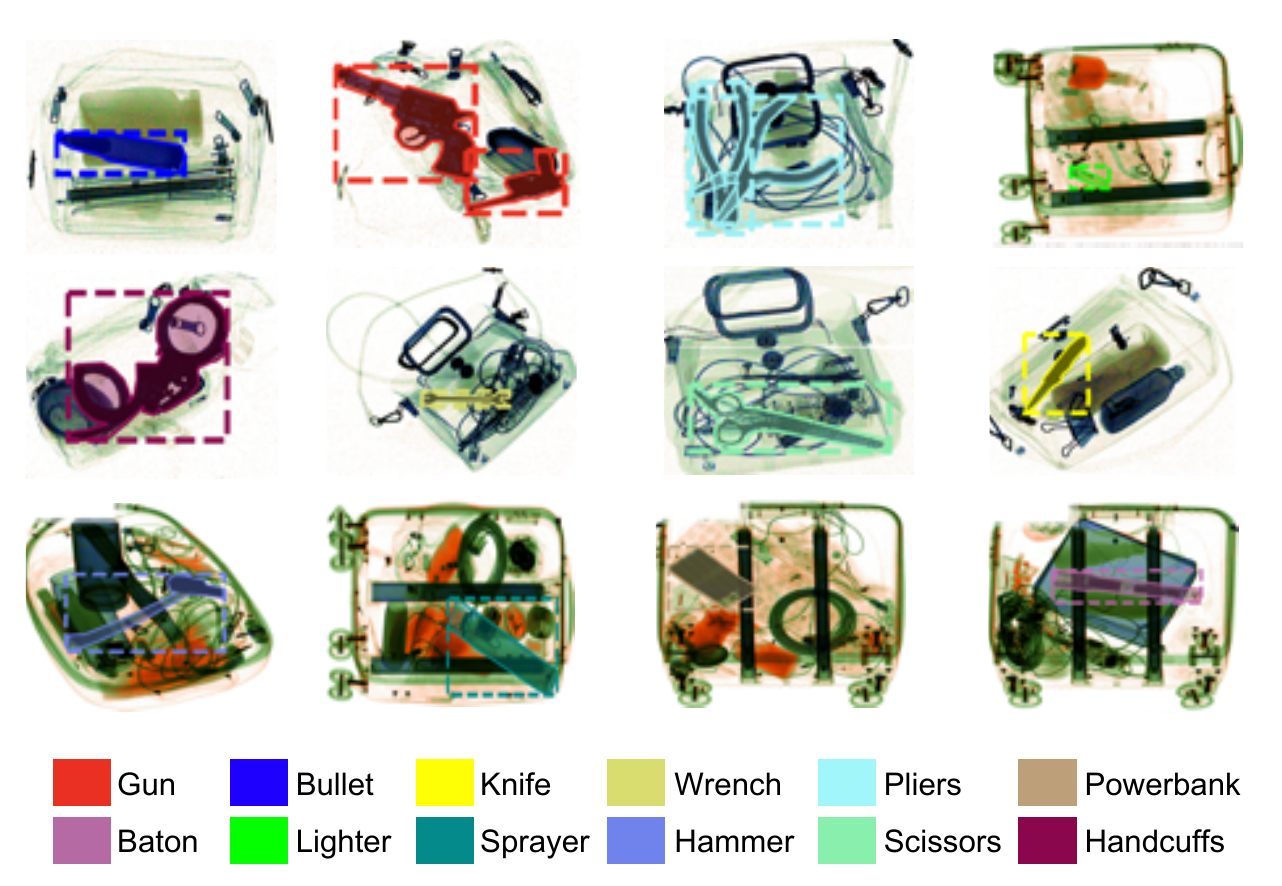}
\caption{Example images in the PIDray dataset with $12$ categories of prohibited items. Each image is provided with image-level and instance-level annotation. For clarity, we show one category per image.}
\label{fig:samples}
\end{figure}

To that end, we collect a large-scale prohibited item detection dataset (PIDray) towards real-world applications. Our PIDray dataset covers $12$ common prohibited items in X-ray images. Some example images with annotations are shown in Figure \ref{fig:samples}, where each image contains at least one prohibited item with both the bounding box and mask annotations. Notably, for better usage, the test set is divided into three subsets, \ie, \textit{easy}, \textit{hard} and \textit{hidden}. The \textit{hidden} subset focuses on the prohibited items deliberately hidden in messy objects (\eg, change the item shape by wrapping wires). To the best of our knowledge, it is the largest dataset for the detection of prohibited items to date. 

Meanwhile, we also present the selective dense attention network (SDANet) to construct a strong baseline, which consists of two modules, \ie, the dense attention module and the dependency refinement module. The dense attention module uses both the spatial and channel-wise attention mechanisms to exploit discriminative features, which is effective to locate the deliberately prohibited items hidden in messy objects. The dependency refinement module is constructed to exploit the dependencies among multi-scale features. Extensive experiments on the proposed dataset show that our method performs favorably against the state-of-the-art methods. Especially, our SDANet achieves $1.5\%$ and $1.3\%$ AP improvements over Cascade Mask R-CNN \cite{DBLP:journals/corr/abs-1906-09756} for object detection and instance segmentation on the {\em hidden} subset, respectively.

The main contributions of this work are summarized as follows. (1) Towards the prohibited item detection in real-world scenarios, we present a large-scale benchmark, \ie, PIDray, formed by $47,677$ images in total. To the best of our knowledge, it is the largest X-ray prohibited item detection dataset to date. Meanwhile, it is the first benchmark aiming at cases where the prohibited items are deliberately hidden in messy objects. (2) We propose the selective dense attention network, formed by the dense attention module and the dependency refinement module. The dense attention module is used to capture the discriminative features in both spatial and channel-wise, and the dependency refinement module is constructed to exploit the dependencies among multi-scale features. (3) Extensive experiments are conducted on the proposed dataset to verify the effectiveness of the proposed method compared to the state-of-the-art methods.

\section{Related Work}
\label{section:related work}

\begin{table*}[t]
  \centering
  \caption{Comparison of the dataset statistics with existing X-ray benchmarks. ``Total'' and ``Prohibited'' indicate the number of total images and the images containing prohibited items in the dataset, respectively. \textbf{C}, \textbf{O}, and \textbf{I} represent Classification, Object Detection, and Instance Segmentation respectively. \textbf{S}, \textbf{A}, and \textbf{R} represent Subway, Airport, and Railway Station respectively.}
  \vspace{4pt}
  \label{table:comparison}
  \setlength{\tabcolsep}{1.0mm}{
    \begin{tabular}{c|c|c|c|c|c|c|c|c|c|c|c}
    \hline
    \multirow{2}[4]{*}{Dataset} & \multirow{2}[4]{*}{Year} & \multirow{2}[4]{*}{Classes} & \multicolumn{2}{c|}{Images} & \multicolumn{3}{c|}{Annotations} & \multirow{2}[4]{*}{Type} & \multirow{2}[4]{*}{Scene} & \multirow{2}[4]{*}{Application} & \multirow{2}[4]{*}{Availability} \\
\cline{4-8}          &       &       & Total & Prohibited & Image & Bbox  & Mask  &       &       &       &  \\
    \hline
    GDXray \cite{mery2015gdxray} & $2015$  & $3$     & $8,150$ & $8,150$ & \checkmark     & \checkmark     &       & Real  & -     & C+O   & \checkmark \\
    Dbf$_{6}$ \cite{akcay2017evaluation}  & $2017$  & $6$     & $11,627$ & $11,627$ & \checkmark     & \checkmark     &       & Real  & -     & C+O   &  $\times$\\
    Dbf$_{3}$ \cite{akcay2018using}  & $2018$  & $3$     & $7,603$ & $7,603$ & \checkmark     & \checkmark     &       & Real  & -     & C+O   &  $\times$\\
    Liu~\etal~\cite{Jinyi2019xray}    & $2019$  & $6$     & $32,253$ & $12,683$ & \checkmark     & \checkmark     &       & Real  & S     & C+O   &  $\times$\\
    SIXray \cite{miao2019sixray} & $2019$  & $6$     & $\bf1,059,231$ & $8,929$ & \checkmark     & \checkmark     &       & Real  & S     & C+O   & \checkmark \\
    OPIXray \cite{wei2020occluded} & $2020$  & $5$     & $8,885$ & $8,885$ & \checkmark     & \checkmark     &       & Synthetic & A     & C+O   & \checkmark \\
    \hline
    \hline
    Ours  & $2021$  & $\bf12$    & $47,677$ & $\bf47,677$ & \checkmark     & \checkmark     & \checkmark     & Real  & S+A+R &    C+O+I   & \checkmark \\
    \hline
    \end{tabular}}%
\end{table*}%

\subsection{Prohibited Items Benchmarks}
When the X-ray passes through an object, different materials absorb the X-ray to different degrees due to its strong penetrating power. Therefore, different materials show different colors in X-ray images. This ability makes it difficult to detect overlapping data. In addition, the difficulties caused by natural images still exist, including intra-class differences, data imbalance, and occlusion.

To advance robust prohibited item detection methods, previous works collect a few datasets. \cite{mery2015gdxray} propose a public dataset called GDXray for nondestructive testing. GDXray contains three types of prohibited items: \textit{gun}, \textit{shuriken} and \textit{razor blade}. Since there is almost no complex background and overlap, it is easy to recognize or detect objects in this dataset. Compared with GDXray, Dbf6 \cite{akcay2017evaluation}, Dbf3 \cite{akcay2018using} and OPIXray \cite{wei2020occluded} contain complicated background and overlapping-data, but the number of images and the number of prohibited items are still insufficient. Recently, \cite{Jinyi2019xray} construct a dataset containing $32,253$ X-ray images, of which $12,683$ images include prohibited items. This dataset contains $6$ types of items, but none of them are strictly prohibited, such as \textit{mobile phones}, \textit{umbrellas}, \textit{computers}, and \textit{keys}. \cite{miao2019sixray} release a large-scale security inspection benchmark named as SIXray, which contains $1,059,231$ X-ray images with image-level annotation. However, fewer images contain prohibited items in the dataset (\ie, only $0.84\%$). In addition, the dataset contains $6$ categories of prohibited items, but only $5$ categories are actually annotated. Different from the aforementioned datasets, we propose a new large-scale security inspection benchmark that contains over $47k$ images with prohibited items and $12$ categories of prohibited items with pixel-level annotation. Towards real-world application, we focus on detecting deliberately hidden prohibited items.


\subsection{Object Detection}
Object detection is one of the fundamental tasks in the computer vision community. Modern object detectors are generally divided into two groups: two-stage and one-stage detectors. 

\textbf{Two-stage Detectors.} R-CNN \cite{girshick2014rich} is one of the first works to show that CNN can dramatically improve the detection performance. However, each regional proposal is processed separately in RCNN, which is very time-consuming. Fast-RCNN\cite{girshick2015fast} proposes the ROI pooling layer, which can extract fixed-size features for each proposal from the feature map of the full image. Faster R-CNN \cite{ren2015faster} introduces the RPN network to replace selective search, which inspires a lot of later work. For example, FPN \cite{lin2017feature} combines low-resolution features with high-resolution features through a top-down pathway and lateral connections. Mask R-CNN \cite{he2017mask} adds a mask branch on the basis of Faster-RCNN\cite{ren2015faster} to improve the detection performance through multi-task learning. Cascade R-CNN \cite{cai2018cascade} applies the classic cascade architecture to Faster R-CNN \cite{ren2015faster}. Libra R-CNN \cite{pang2019libra} develops a simple and effective framework to eliminate the imbalance in the detection training process.

\textbf{One-stage Detectors.} OverFeat \cite{sermanet2013overfeat} is one of the first deep learning based one-stage detectors. After that, different one-stage object detectors are proposed, including SSD \cite{liu2016ssd}, DSSD \cite{fu2017dssd}, and YOLO series \cite{redmon2016you,redmon2017yolo9000,redmon2018yolov3}. RetinaNet \cite{lin2017focal} greatly improves the accuracy of one-stage detector, making it possible for one-stage detector to surpass two-stage detector. Recently, anchor-free approaches have attracted wide attention of researchers by using key points to represent the objects, including CornerNet \cite{law2018cornernet}, CenterNet \cite{duan2019centernet}, and FCOS \cite{tian2019fcos}. These methods eliminate the need for anchors and provide a simplified detection framework. 

\subsection{Attention Mechanism}
Recently, attention mechanism has been widely used in a variety of tasks, such as neural machine translation, image captioning, and visual question answering. The essence of the attention mechanism is to imitate human visual attention, which can quickly filter out discriminative information from a large number of information. In order to obtain more discriminative information, various attention mechanisms have been proposed. SENet \cite{hu2018squeeze} proposes the Squeeze-and-Excitation module to model the interdependence between channels. CBAM \cite{woo2018cbam} models the inter-channel relation and the inter-spatial relation of features. Non-Local network \cite{wang2018non} can capture the remote dependency of any two locations directly, which calculates the weighted sum of the features of all positions in the input feature map as the response of a certain position. As many previous works \cite{lin2017feature,liu2018path} show the importance of multi-scale feature fusion, we think it is the key technology to solve the problem of prohibited item detection. In X-ray images, many important details of objects are missing, such as texture and appearance information. Moreover, the contours of objects overlap, which also brings great challenges to detection. Multi-scale feature fusion considers the low-level layers with rich detail information and the high-level layers with rich semantic information, which can better detect the prohibited item. Therefore, we propose a selective dense attention network. Specifically, we learn the relations between feature maps across different stages at inter-channel and inter-pixel positions.

\begin{figure}[t]
\centering
\includegraphics[width=1.0 \linewidth]{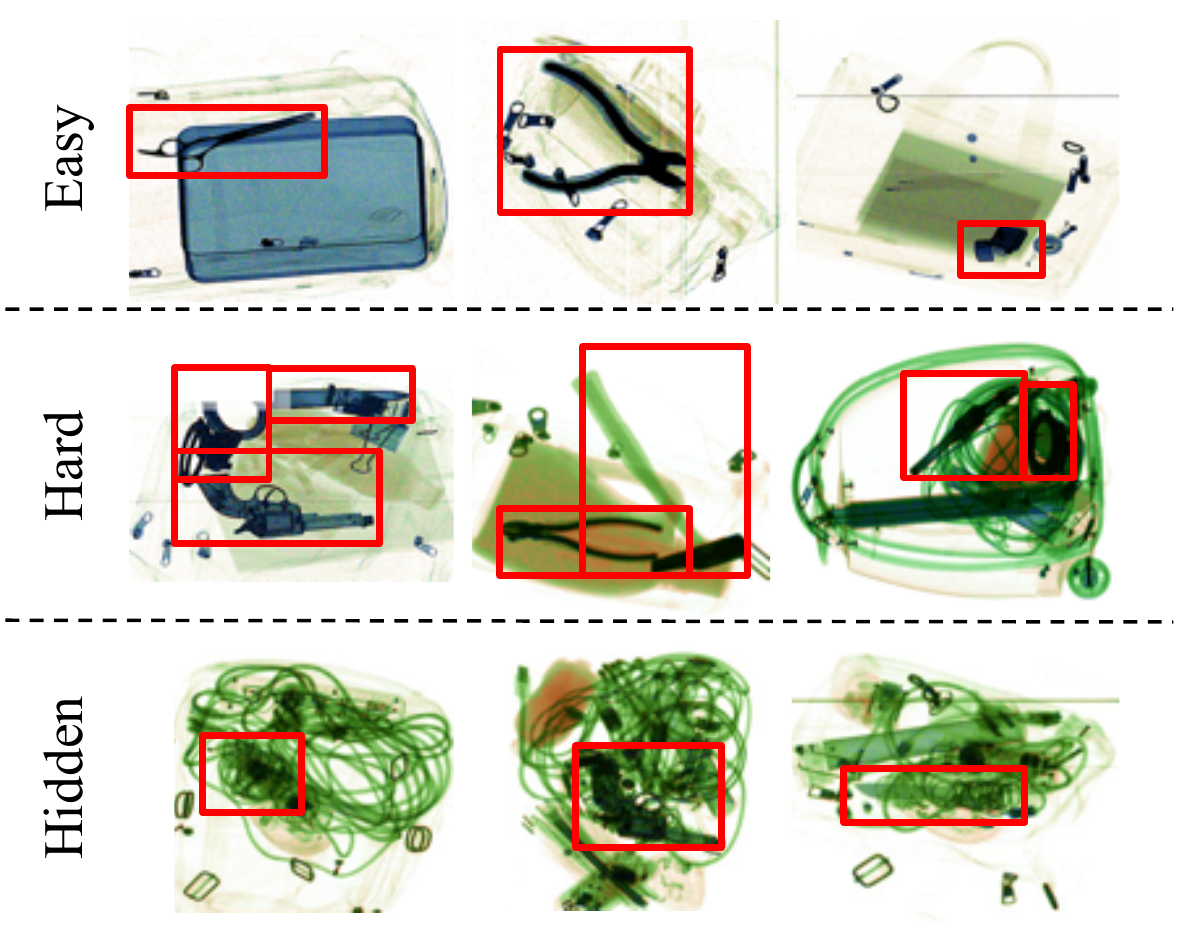}
\caption{Examples of test sets with different difficulty levels in the proposed PIDray dataset. From top to bottom, the degree of difficulty gradually increases. }
\label{fig:testset_sample}
\end{figure}

\section{The PIDray Dataset}
In this section, we provide details of the constructed PIDray dataset, including the data collection, annotation, and statistical information.

\subsection{Data Collection}
The PIDray dataset was collected in different scenarios (such as airports, subway stations, and railway stations), where we were allowed to place a security inspection machine. We recruited volunteers who did not mind displaying their packages in the dataset (we promise to use it only for scientific research and not for business). We use 3 security inspection machines from different manufacturers to collect X-ray data. Images generated by different machines have certain differences in the size and color of the objects and background. After sending the package to the security inspection machine, the machine will completely cut out the package by detecting the blank part of the image. Generally speaking, the image height is fixed while the image width relies on the size of the package being scanned.

\begin{figure}[t]
\centering
\includegraphics[width=0.99\linewidth]{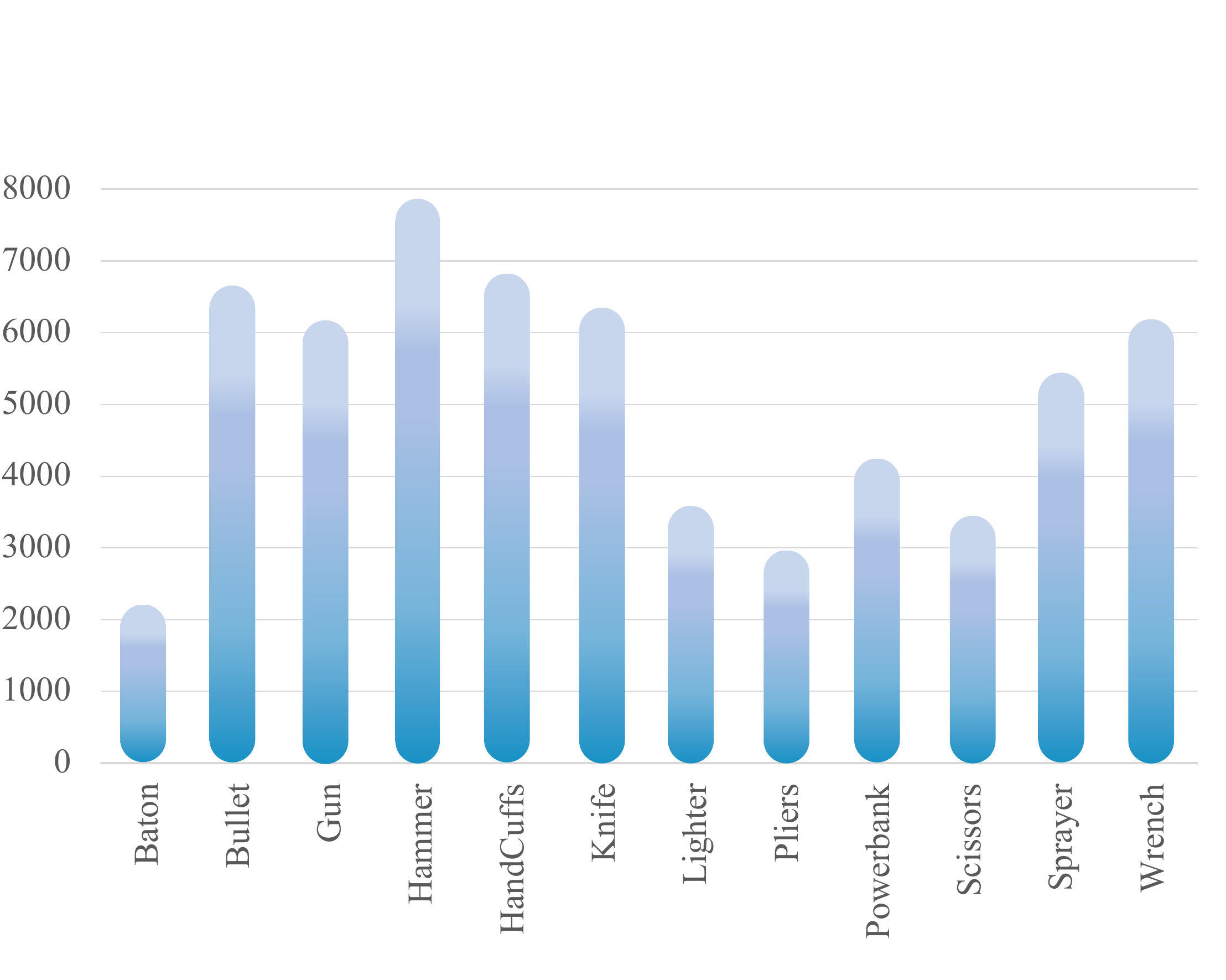}
\caption{Class distribution of the PIDray dataset. The blue bar represents the number of each class in the PIDray dataset.}
\label{fig:distribute}
\end{figure}

The complete collection process is as follows: when the person is required for security inspection, we randomly put the pre-prepared prohibited items in the package he or she is carrying. At the same time, the rough area of the object was saved, so that the subsequent annotation work can be carried out smoothly. There are a total of $12$ categories of prohibited items defined in the dataset, namely \textit{gun}, \textit{knife}, \textit{wrench}, \textit{pliers}, \textit{scissors}, \textit{hammer}, \textit{handcuffs}, \textit{baton}, \textit{sprayer}, \textit{powerbank}, \textit{lighter} and \textit{bullet}. To keep diversity, we prepare $2\sim15$ instances for every kind of prohibited item. We spend more than three months collecting a total of $47,677$ images for the PIDray dataset. Finally, the distribution of each category in the dataset is summarized in Figure \ref{fig:distribute}. All images are stored in PNG format. 

\begin{table}[t]
  \centering
  \small
  \fontsize{10}{8}\selectfont
  \caption{Statistics of the PIDray dataset.}
  \vspace{4pt}
  \label{table:partition_dataset}
  \setlength{\tabcolsep}{3.7mm}{
    \begin{tabular}{ccccc}
    \toprule
    \multirow{2}[3]{*}{Mode} & \multirow{2}[3]{*}{Train} & \multicolumn{3}{c}{Test} \\
\cmidrule{3-5}          &       & Easy  & Hard  & Hidden \\
    \midrule
    Count & $29,457$ & $9,482$  & $3,733$  & $5,005$ \\
    \midrule
    Total & \multicolumn{4}{c}{$47,677$} \\
    \bottomrule
    \end{tabular}}%
    
\end{table}%

\begin{figure*}[t]
\centering
\includegraphics[width=0.95\linewidth]{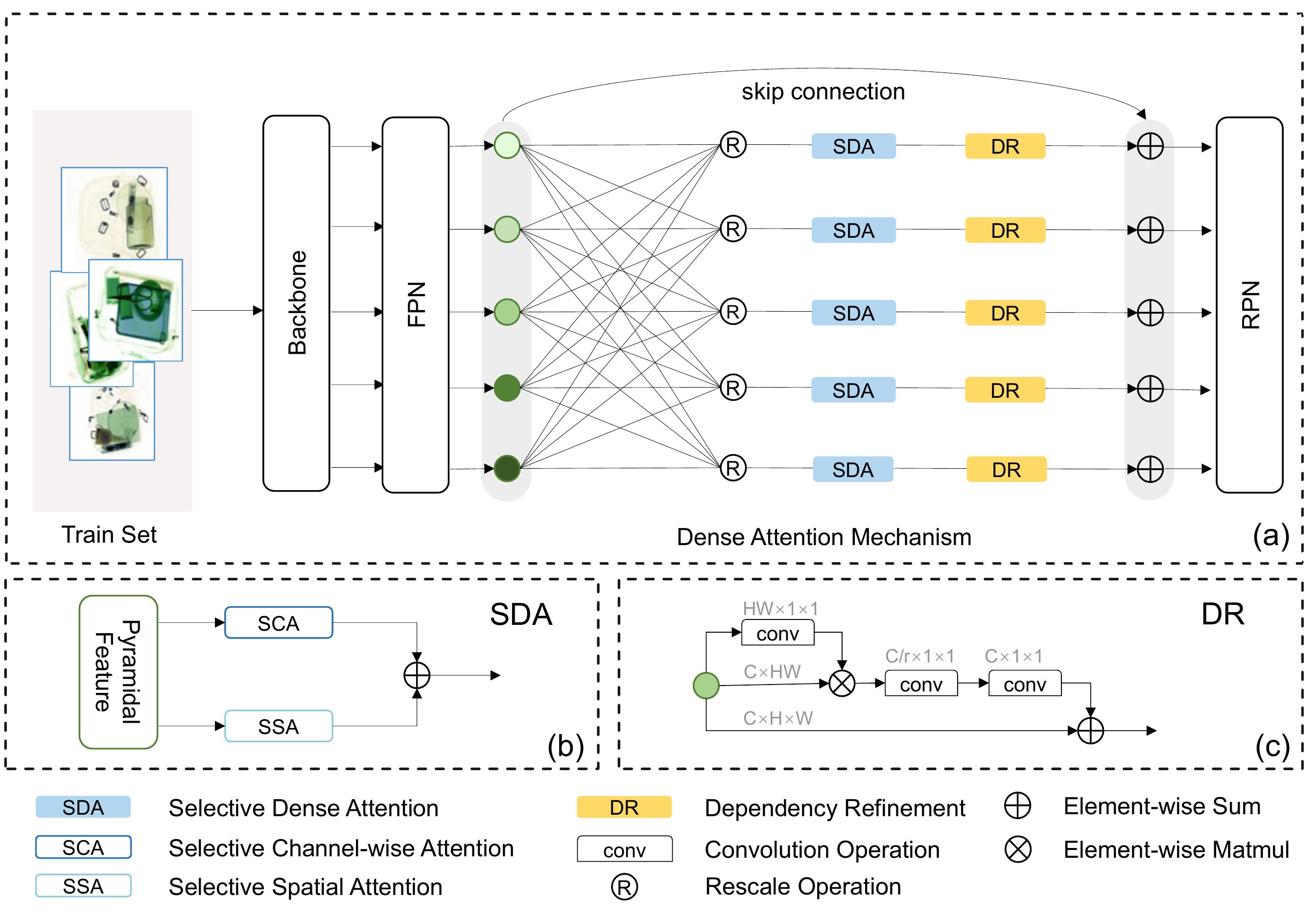}
\caption{Network architecture. (a) The overall architecture of the proposed selective dense attention network. (b) The selective dense attention module. (c) The dependency refinement module.}
\label{fig:structure}
\end{figure*}

\subsection{Data Annotation}
We recruited some volunteers to annotate the collected data. In order to enable them to identify prohibited items from X-ray images more quickly and accurately, some training courses have been organized. We first organized 5 volunteers to filter out images from the dataset that contain no prohibited items. At the same time, they also need to annotate the image-level labels, which can facilitate the later annotation work. In terms of annotation, we organized over $10$ volunteers to label our dataset using the labelme tool\footnote{\url{http://labelme.csail.mit.edu/Release3.0/}} for two months. Each image takes about $3$ minutes to annotate, and each volunteer spends about $10$ hours to annotate the image every day. During the annotation process, we label both the bounding box and the segmentation mask of each instance. After multiple rounds of double-check, the errors are minimized as much as possible. Finally, we generate high-quality annotations for each image.

\subsection{Data Statistics}
As far as we know, the PIDray dataset is the largest X-ray prohibited item detection dataset to date. It contains 47,677 images and 12 classes of prohibited items. As presented in Table \ref{table:partition_dataset}, we split those images into 29,457(roughly 60\%) and 18,220(remaining 40\%) images as training and test sets, respectively. In addition, according to the difficulty degree of prohibited item detection, we group the test set into three subsets, \ie, \textit{easy}, \textit{hard} and \textit{hidden}. Specifically, the \textit{easy} mode means that the image in the test set contains only one prohibited item. The \textit{hard} mode indicates that the image in the test set contains more than one prohibited item. The \textit{hidden} mode indicates that the image in the test set contains deliberately hidden prohibited items. As shown in Figure \ref{fig:testset_sample}, we provide several examples in the test set with different difficulty levels. 

\section{Selective Dense Attention Network}
As discussed above, the previous works usually employ feature pyramid \cite{lin2017feature} to exploit multi-scale feature maps in the network, which focuses on fusing features only in adjacent layers. After that, the succinct heads (\eg, a simple convolutional layer) are applied on the pooled feature grid to predict bounding boxes and masks of instances. However, the performance suffers from scale variation of objects in complex scenes. Our goal is to learn the importance of multi-scale feature maps based on top-down feature pyramid structure \cite{lin2017feature}. In this section, we will introduce the architecture and components of the proposed Selective Dense Attention Network (SDANet) in detail.

\subsection{Network Architecture}
As shown in Figure \ref{fig:structure}(a), following the feature pyramid, our network further makes full use of multi-scale feature maps by the following two critical steps: 1) Fusing information from different layers by two selective attention modules. 2) Enhancing the fused features by the dependency refinement module. 

Note that the two steps are performed on the feature map in each layer. After combining both the original and enhanced maps, the multi-scale representation is fed into the Region Proposal Network (RPN) for final prediction.

Inspired by the work of \cite{DBLP:conf/cvpr/LiW0019}, we propose two selective attention modules to extract channel-wise and spatial attention of different feature maps in the pyramid respectively, including the Selective Channel-wise Attention module (SCA) and the Selective Spatial Attention module (SSA). As shown in Figure \ref{fig:structure}(b), each feature map in the pyramid is fed into SCA and SSA respectively. At the $i$-th layer, the output enhanced feature is calculated by element-wise summation of features after the two modules. To implement the SCA and SSA modules, we first fuse features in different layers through element-wise operations, \ie, $\hat{X} = \sum_{i=1}^{n}X_{i}$. Thus we achieve a global semantic representation among different maps. Note that, we resize the multi-stage features $\left\{X_{1},\cdots,X_{n}\right\}$ to the same scale as the $i$-th layer feature before feeding them into the two modules. Then, we obtain enhanced features by aggregating feature maps with various attentions, which is described in detail as follows.

\subsection{Selective Channel-wise Attention}
\begin{figure}[t]
\centering
\includegraphics[width=1.0\linewidth]{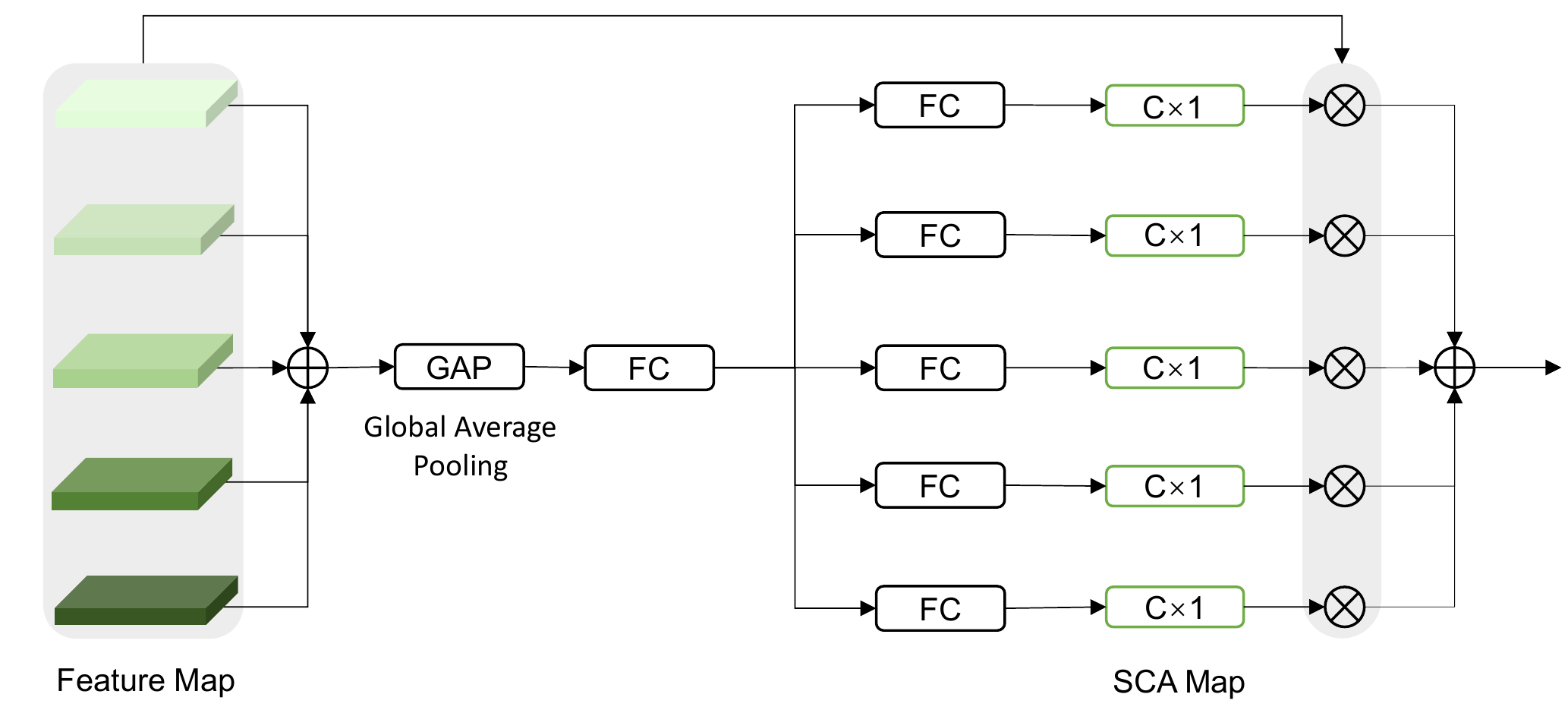}
\caption{Illustration of selective channel-wise attention module (SCA).}
\label{figure:channel_attention}
\end{figure}

As shown in Figure \ref{figure:channel_attention}, we employ the global average pooling (GAP) layer to obtain global channel information based on the base feature $\hat{X}$. After that, we use the fully connected (FC) layer to squeeze global channel information by reducing the channel dimension (\eg, from $256$ to $128$). Further, we obtain the channel-wise attention weights $\{\omega^{c}_{i}\}_{i=1}^n$ of different feature maps adaptively by adding FC layers and softmax operation for each layer. Finally, the enhanced feature map $V_{C}$ is obtained by the attention weight on each layer, \ie, $V_{C} = \sum_{i=1}^{n}\omega^{c}_{i}\cdot X_{i}$.

\subsection{Selective Spatial Attention}
\begin{figure}[t]
\centering
\includegraphics[width=1.0\linewidth]{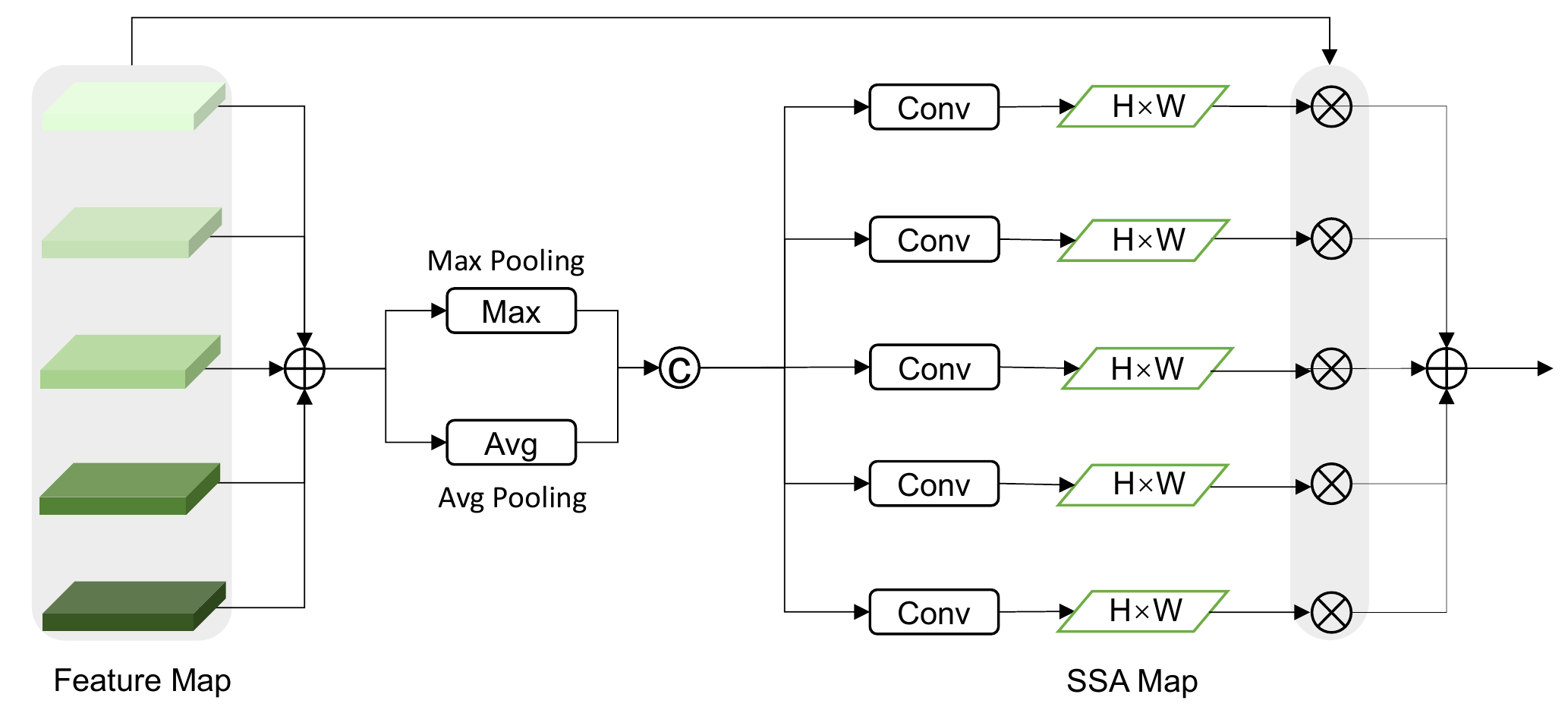}
\caption{Illustration of selective spatial attention module (SSA).}
\label{figure:spatial_attention}
\end{figure}

As shown in Figure \ref{figure:spatial_attention}, we use both the average pooling and maximum pooling operations on the feature map $\hat{X}$ to generate two different spatial context descriptors, \ie, $\text{Avg}(\hat{X})$, $\text{Max}(\hat{X})$. Given the concatenated context descriptors, we can obtain the spatial attention weights by adding convolutional layers and softmax operation for each layer. Finally, the feature map $V_{S}$ is obtained by the attention weight on each layer, \ie, $V_{S}(x,y) = \sum_{i=1}^{n}\omega^{s}_{i}(x,y)\cdot X_{i}(x,y)$, where $(x,y)$ indicates the index of pixel in feature map.

\subsection{Dependency Refinement}
After obtaining the aggregated features with both channel and spatial attention, we develop the Dependency Refinement (DR) module to generate more discriminative feature maps. Non-local representation \cite{wang2018non} can capture long-range dependencies effectively, which further improves the accuracy. As shown in Figure \ref{fig:structure}(c), we first aggregate global context features, and then establish relationships between different channels. Finally, the global context feature is merged into features of all positions by a fusion module.

\begin{table*}[t]
  \centering
  \small
  \caption{The evaluation results on the proposed PIDray dataset. COCO mmAP (\%)  is used to evaluate performance of all methods.}
  \vspace{4pt}
  \setlength{\tabcolsep}{2.85mm}
    \begin{tabular}{c|c|cccc|cccc}
    \hline
    \multirow{2}{*}{Method} & \multirow{2}{*}{Backbone} &  \multicolumn{4}{c|}{Detection AP}     & \multicolumn{4}{c}{Segmentation AP} \\
\cline{3-10}          &       & Easy  & Hard  & Hidden & Overall & Easy  & Hard  & Hidden  & Overall \\
    \hline
    FCOS  & ResNet-101-FPN  &  $61.8$  &  $51.7$  &  $37.5$  &  $50.3$  & -     & -     & -     & - \\
    RetinaNet  & ResNet-101-FPN  &  $61.8$  &  $52.2$  &  $40.6$  &  $51.5$  & -     & -     & -     & - \\
    Faster R-CNN & ResNet-101-FPN  & $63.3$  & $57.2$  & $42.1$  & $54.2$  & -     & -     & -     & - \\
    Libra R-CNN & ResNet-101-FPN  & $64.7$  & $58.8$  & $42.9$  & $55.5$  & -     & -     & -     & - \\
    Mask R-CNN & ResNet-101-FPN  & $64.7$  & $59.0$    & $43.8$  & $55.8$  & $57.6$  & $50.2$  & $35.2$  & $47.7$  \\
    SSD512 & VGG16  & $68.1$  & $58.9$  & $45.7$  & $57.6$  & -     & -     & -     & - \\
    Cascade R-CNN & ResNet-101-FPN  & $69.3$  & $62.8$  & $48.0$    & $60.0$  & -     & -     & -     & - \\
    Cascade Mask R-CNN & ResNet-101-FPN  & $70.9$  & $64.0$    & $48.0$    & $61.0$  & $59.2$  & $51.5$  & $36.1$  & $48.9$  \\
    SDANet(ours) & ResNet-101-FPN & $\textbf{71.2}$ & $\textbf{64.2}$ & $\textbf{49.5}$ & $\textbf{61.6}$  & $\textbf{59.9}$ & $\textbf{52.0}$ & $\textbf{37.4}$ & $\textbf{49.8}$  \\
    \hline
    \hline
    Cascade Mask R-CNN & ResNet-101-BiFPN  & $68.0$  & $61.1$    & $46.9$    & $58.7$  & $58.0$  & $49.8$  & $35.3$  & $47.7$  \\
    Cascade Mask R-CNN & ResNet-101-PAFPN  & $70.4$  & $63.4$    & $46.7$    & $60.2$  & $59.2$  & $51.4$  & $35.0$  & $48.5$  \\
    Cascade Mask R-CNN & ResNet-101-FPN  & $70.9$  & $64.0$    & $48.0$    & $61.0$  & $59.2$  & $51.5$  & $36.1$  & $48.9$  \\
    SDANet(ours) & ResNet-101-FPN & $\textbf{71.2}$ & $\textbf{64.2}$ & $\textbf{49.5}$ & $\textbf{61.6}$  & $\textbf{59.9}$ & $\textbf{52.0}$ & $\textbf{37.4}$ & $\textbf{49.8}$  \\
    \hline
    \end{tabular}%
  \label{table:baseline_results}%
\end{table*}%

\begin{table*}[t]
  \centering
  \small
  \caption{Effectiveness of various designs. All models are trained on the PIDray {\it training} subset and tested on the PIDray hidden {\it test} set. The accuracies are indicated by ``detection AP/segmentation AP''.}
  \vspace{4pt}
  \setlength{\tabcolsep}{2.4mm}{
\begin{tabular}{ccc|cccccccc}
    \hline
    SCA & SSA & DR& AP    & $\text{AP}_{50}$  & $\text{AP}_{75}$  & $\text{AP}_{S}$   & $\text{AR}_{1}$   & $\text{AR}_{10}$  & $\text{AR}_{100}$ & $\text{AR}_{S}$ \\
    \hline
          &       &       &       $48.0/36.1$    & $62.7/58.9$  & $54.0/40.4$    & $57.0/43.5$    & $56.0/42.9$    & $57.6/44.0$  & $57.6/44.0$  & $57.6/44.0$ \\
    \checkmark     &       &             & $48.3/36.5$  & $63.5/59.3$  & $54.3/41.2$  & $57.2/43.9$  & $56.2/43.4$  & $57.9/44.4$  & $57.9/44.4$  & $57.9/44.4$ \\
         &    \checkmark   &            & $48.3/36.2$  & $63.2/59.6$  & $54.6/40.1$  & $57.4/43.8$  & $56.6/43.3$  & $58.1/44.3$  & $58.1/44.3$  & $58.1/44.3$ \\
    \checkmark     & \checkmark     &            & $48.9/36.7$  & $63.8/60.0$  & $55.4/40.8$  & $58.3/44.3$  & $57.4/43.8$  & $59.3/45.0$  & $59.3/45.0$  & $59.3/45.0$ \\
    \checkmark     & \checkmark     & \checkmark        & $49.5/37.4$  & $64.5/60.6$  & $55.7/42.2$  & $58.5/44.8$  & $57.2/44.1$  & $59.5/45.5$  & $59.5/45.5$    & $59.5/45.5$ \\
    \hline
    \end{tabular}}
  \label{table:ablation_study}%
\end{table*}%

\section{Experiment}
We conduct extensive experiments on the PIDray dataset to compare the proposed method with several state-of-the-art algorithms. Then, the ablation study is used to show the effectiveness of the proposed modules in our method. Finally, we verify the effectiveness of the proposed method on general detection datasets.
\subsection{Implementation Details}
We employ the MMDetection toolkit\footnote{\url{https://github.com/open-mmlab/mmdetection}} to implement our method, which is performed on a machine with two NVIDIA Tesla V100 cards. Our method is implemented in Pytorch. For a fair comparison, all the compared methods are trained on the training set and evaluated on the test set of the PIDray dataset. The proposed SDANet is based on Cascade Mask-RCNN \cite{DBLP:journals/corr/abs-1906-09756}, where the ResNet-101 network is used as the backbone. According to our statistics, the average resolution of the images in our dataset is approximately $500 \times 500$. Therefore, we resize the image to $500 \times 500$ for compared detectors for a fair comparison. The entire network is trained with a stochastic gradient descent (SGD) algorithm with a momentum of $0.9$ and a weight decay of $0.0001$. The initial learning rate is set as $0.02$ and the batch size is set as $2$. Unless otherwise specified, other parameters involved in the experiment follow the settings of MMdetection.

\subsection{Evaluation Metrics}
According to the evaluation metric of MS COCO\cite{lin2014microsoft}, we evaluate the performance of the compared methods on our PIDray dataset using both the AP and AR metrics. The scores are averaged over multiple Intersection over Union (IoU). Notably, we use $10$ IoU thresholds between $0.50$ and $0.95$. Specifically, the AP score is averaged across all $10$ IoU thresholds and all $12$ categories. In order to better assess a model, we look at various data splits. $\text{AP}_{50}$ and $\text{AP}_{75}$ scores are calculated at IoU $=0.50$ and IoU $=0.75$ respectively. Note that many prohibited items are small (area $<32^2$) in the PIDray dataset, which is evaluated by the $\text{AR}_{S}$ metric. Besides, the $\text{AR}$ score is the maximum recall given a fixed number of detections (\eg, $1,10,100$) per image, averaged over $12$ categories and $10$ IoUs. 

\begin{figure*}[t]
\centering
\includegraphics[width=1.0\linewidth]{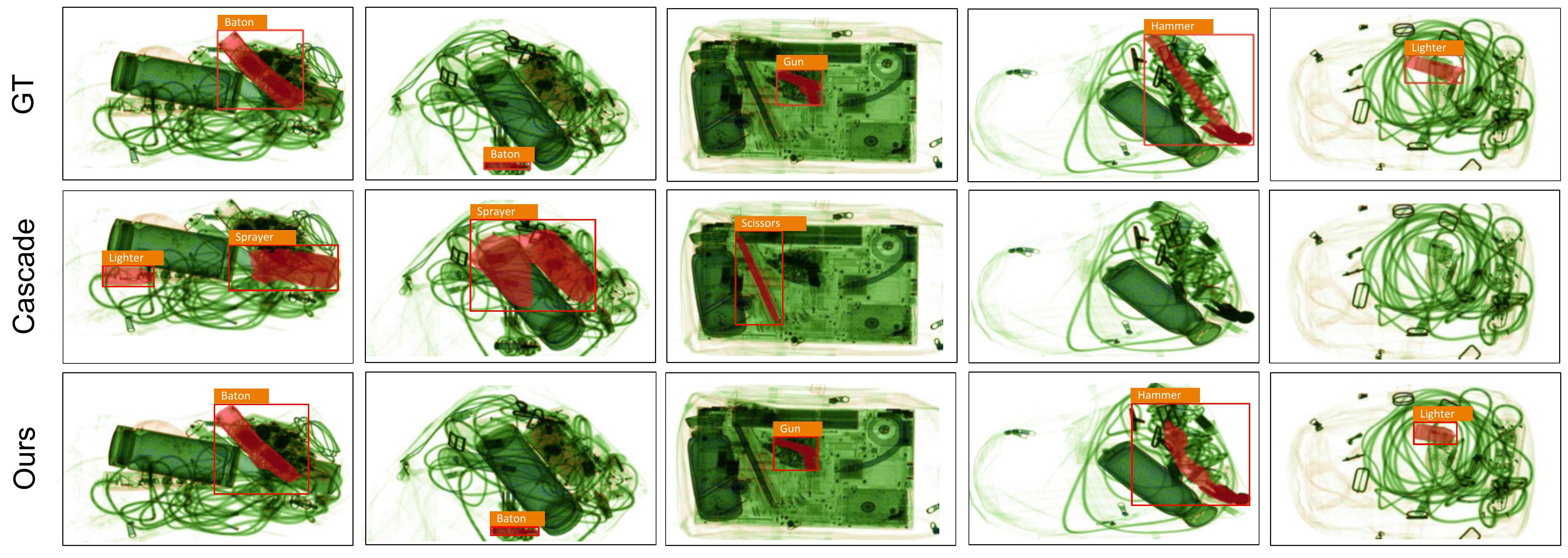}
\caption{Comparison between the proposed SDANet method and Cascade Mask R-CNN \cite{DBLP:journals/corr/abs-1906-09756}. GT indicates Ground-truth, Cascade indicates the results generated by Cascade Mask R-CNN, and Ours indicates the results generated by SDANet.}
\label{figure:visual_results}
\end{figure*}

\subsection{Overall Evaluation}
As presented in Table \ref{table:baseline_results}, we firstly compare our method with a few state-of-the-art object detectors. It can be seen that our SDANet achieves the best performance in terms of all the subsets in the PIDray dataset. For example, compared with the biggest competitor Cascade Mask R-CNN \cite{DBLP:journals/corr/abs-1906-09756}, our method achieves $1.5\%$ and $1.3\%$ AP gain for the two sub-tasks on the hidden test set, which shows the effectiveness of the proposed selective dense attention module. As shown in Figure \ref{figure:visual_results}, our method achieves higher accuracy than Cascade Mask R-CNN \cite{DBLP:journals/corr/abs-1906-09756}. The visual results show that SDANet can effectively detect prohibited items, especially those that have been deliberately hidden.

To verify the effectiveness of the proposed selective dense attention scheme, we compare our method with the previous multi-scale feature fusion strategies including FPN \cite{lin2017feature}, PAFPN \cite{DBLP:conf/cvpr/LiuQQSJ18}, and BiFPN \cite{DBLP:conf/cvpr/TanPL20}. FPN \cite{lin2017feature} provides a top-down pathway to fuse multi-scale features, while PAFPN \cite{DBLP:conf/cvpr/LiuQQSJ18} adds an additional bottom-up pathway on top of FPN. BiFPN \cite{DBLP:conf/cvpr/TanPL20} is weighted bi-directional feature pyramid network, which allows easy and fast multi-scale feature fusion. As presented in Table \ref{table:baseline_results}, our method outperforms existing multi-scale feature fusion strategies. We speculate that this is attributed to two reasons. First, two selective attention modules can aggregate semantic information across multi-layers densely. Second, the dependency refinement module can further capture long-range dependencies among different feature maps. The results indicate that our method can detect deliberately hidden data effectively. 

\subsection{Ablation Study}
Since this work focuses on detecting prohibited items that are hidden deliberately, we conduct the ablation study to analyze the influence of the proposed modules on the hidden test set of the PIDray dataset. 

As presented in Table \ref{table:ablation_study}, we report how the performance of our SDANet is improved when we add the module one by one in the baseline Cascade Mask R-CNN \cite{DBLP:journals/corr/abs-1906-09756}. Firstly, the selective channel-wise attention module improves the baseline method by $0.3\%$ detection AP and $0.4\%$ segmentation AP. Then, the performance continuously improves by $0.6\%$ detection AP and $0.2\%$ segmentation AP when incorporating the selective spatial attention modules. Finally, the dependency refinement module contributes to a $0.6\%$ and $0.7\%$ improvement in terms of detection AP and segmentation AP, respectively. 

We also compare the dependency refinement module with the existing attention mechanisms(e.g. SE and CBAM). Table \ref{table:DR_comparison} shows the results of all models. The results show that DR has obvious advantages in detecting deliberately hidden items.
\begin{table}[t]
\setlength{\belowcaptionskip}{-0cm}
  \centering
  \caption{Comparison of dependency refinement (DR) and other attention mechanisms on the hidden test set.}
  \vspace{4pt}
    \setlength{\tabcolsep}{7.3mm}{
    \begin{tabular}{c|cc}
    \hline
    Method & Det AP & Seg AP \\
    \hline
    ours w/o DR & 48.9  & 36.7 \\
    \hline
    +SE & 49.1  & 36.7 \\
    +CBAM & 47.0  & 35.8 \\
    +DR   & \textbf{49.5} & \textbf{37.4} \\
    \hline
    \end{tabular}}%
  \label{table:DR_comparison}%
\end{table}%

\begin{table}[t]
  \caption{Evaluation results on the MS COCO and PASCAL VOC detection datasets.}
  \vspace{4pt}
    \centering
    \setlength{\tabcolsep}{5.5mm}{
      \begin{tabular}{c|c|c}
      \hline
      Method & MS COCO  & PASCAL VOC\\
      \hline
      baseline&   42.9    &  81.5\\
      \hline
      SDANet  &   {\bf 43.5}   &  {\bf 82.5} \\
      \hline
      \end{tabular}}%
    \label{tab:general_datasets}%
  \end{table}%
\subsection{Evaluation on General Detection Dataset}
Finally, we also conduct some experiments on general detection datasets to evaluate the effectiveness of SDANet on the natural image. The experiment uses MS COCO\cite{lin2014microsoft} and PASCAL VOC\cite{everingham2010pascal}, which are well-known data sets in the field of natural image detection. The experimental results are shown in Table \ref{tab:general_datasets}. We follow the training and testing pipelines in MMDetection. Compared with the baseline method(Cascade Mask R-CNN), we have achieved 0.6 AP and 1.0 AP gain on MS COCO and PASCAL VOC, respectively. Experimental results demonstrate that our method is not only suitable for the detection of prohibited items, but also effective in general scenarios. 

\section{Conclusion}

In this paper, we construct a challenging dataset(namely PIDray) for prohibited item detection, especially dealing with the cases that the prohibited items are hidden in other objects. PIDray is the largest prohibited items detection dataset so far to our knowledge. Moreover, all images are annotated with bounding boxes and masks of instances. To learn the importance of multi-scale feature maps, we propose the selective dense attention network. The experiment on the PIDray dataset proves the superiority of our method. We hope that the proposed dataset will help the community to establish a unified platform for evaluating the prohibited item detection methods towards real applications. For future work, we plan to extend the current dataset to include more images as well as richer annotations for comprehensive evaluation. 

\section*{Acknowledgement}
We would like to thank Ruyi Ji, Jiaying Li, Xu Wang, and others for their help in data collection and annotation. 
{\small
\bibliographystyle{ieee_fullname}
\bibliography{egbib}
}

\end{document}